\documentclass[conference]{IEEEtran}
\IEEEoverridecommandlockouts

\usepackage{cite}
\usepackage{amsmath,amssymb,amsfonts}
\usepackage{algorithmic}
\usepackage{graphicx}
\usepackage{textcomp}
\usepackage{xcolor}

\usepackage{amsmath,amssymb,amsfonts}
\usepackage{tabularx}  
\usepackage{booktabs} 
\usepackage{array}
\usepackage{multirow}
\usepackage{booktabs}  

\usepackage{xurl}

\def\BibTeX{{\rm B\kern-.05em{\sc i\kern-.025em b}\kern-.08em
    T\kern-.1667em\lower.7ex\hbox{E}\kern-.125emX}}
\begin{document}

\title{LentEx: Generalizable Latent Entity Extraction via Synthetic Data and Instruction-Tuned LLMs\\}


\author{
\IEEEauthorblockN{1\textsuperscript{st} Umesh Bodhwani}
\IEEEauthorblockA{\textit{Amazon} \\
Seattle, WA, USA \\
bodhwani@amazon.com}
\and
\IEEEauthorblockN{2\textsuperscript{nd} Yuan Ling}
\IEEEauthorblockA{\textit{Amazon} \\
Seattle, WA, USA \\
yualing@amazon.com}
\and
\IEEEauthorblockN{3\textsuperscript{rd} Cibi Chakravarthy Senthilkumar}
\IEEEauthorblockA{\textit{Amazon} \\
Seattle, WA, USA \\
sentocb@amazon.com}
\and
\IEEEauthorblockN{4\textsuperscript{th} Shujing Dong}
\IEEEauthorblockA{\textit{Amazon} \\
Seattle, WA, USA \\
shujdong@amazon.com}
\and
\IEEEauthorblockN{5\textsuperscript{th} Yarong Feng}
\IEEEauthorblockA{\textit{Amazon} \\
Seattle, WA, USA \\
yarongf@amazon.com}
\and
\IEEEauthorblockN{6\textsuperscript{th} Hongfei Li}
\IEEEauthorblockA{\textit{Amazon} \\
Seattle, WA, USA \\
lihongfe@amazon.com}
\and
\IEEEauthorblockN{7\textsuperscript{th} Ayush Goyal}
\IEEEauthorblockA{\textit{Amazon} \\
Seattle, WA, USA \\
ayushg@amazon.com}
}


\maketitle

\begin{abstract}
Latent entity extraction (LEE) tackles the challenge of identifying implicit, contextually inferred entities within free text—an area where traditional entity extraction methods fall short. In this paper, we introduce LentEx, a novel framework for latent entity extraction that leverages synthetic data generation and instruction fine-tuning to optimize smaller, efficient large language models (LLMs). Latent entities, which are often abstract and thematic, are crucial for applications such as retrieval-augmented generation (RAG), customer persona analysis, and knowledge graph enrichment. LentEx addresses the scarcity of labeled datasets by employing a template-based approach to generate diverse, contextually rich synthetic data, ensuring high variability and alignment with real-world distributions. To our knowledge, LentEx is the first to systematically approach LEE through the lens of LLMs. LentEx demonstrates significant performance improvements across multiple tasks, notably surpassing state-of-the-art models on the MTEB Clustering Benchmark. Furthermore, our methodology enables robust generalization to unseen domains, making LentEx highly applicable in real-world NLP tasks, including RAG and clustering, thereby establishing a new paradigm for latent entity understanding and extraction in natural language processing.
\end{abstract}


\begin{figure*}[t]
    \centering
    \includegraphics[width=\linewidth]{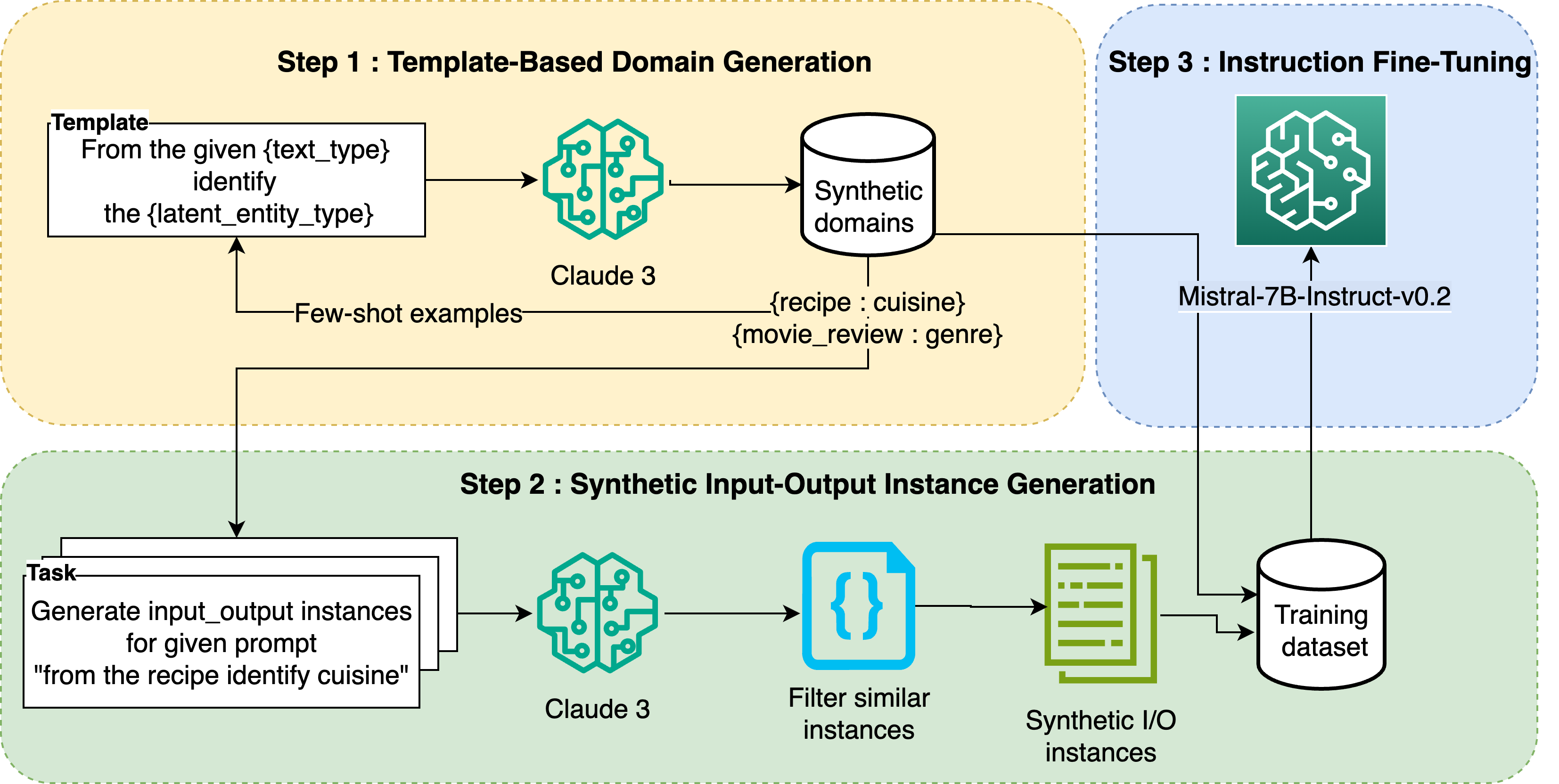} 
    
    \caption{Overview of the LentEx workflow in three steps: (1) Template-based domain generation using Claude-3 with few-shot examples, (2) Synthetic input-output instance generation with similarity filtering, and (3) Instruction fine-tuning using Mistral-7B-Instruct-v0.2}
    \label{fig:example}
\end{figure*}

\section{Introduction}

Latent Entity Extraction (LEE) refers to identifying entities implicitly present in textual data, inferred from contextual nuances rather than explicitly mentioned. Unlike traditional Named Entity Recognition (NER), which focuses on overt entities such as names of organizations, locations, or persons, LEE aims to uncover abstract or thematic entities—such as the persona behind a conversation or an implied industry sector—deduced from context. For instance, while NER may quickly identify mentions of “Google” or “New York,” LEE endeavors to infer broader, latent concepts such as "technology company" or "urban environment" based solely on contextual cues.

Prior research on LEE has predominantly relied on classification or topic modeling. Classification methods are inherently constrained by their reliance on predefined labels, often requiring extensive retraining as new categories emerge \cite{lample-etal-2016-neural,shoshan2018latent}. Topic modeling, although adept at identifying general themes, frequently produces clusters of keywords that lack interpretability and fail to provide actionable insights regarding specific latent entities \cite{blei2003latent, miao2018discovering, grootendorst2022bertopic, Xu_2023_2}. These approaches generate high-level thematic summaries but often lack the granularity to infer nuanced, context-specific information. In retrieval-augmented generation (RAG) systems, LEE enhances document filtering by extracting latent entities from queries and documents. For instance, a query like \textit{What are the latest advancements in AI?} can be enriched with latent entities such as \textit{deep learning techniques} or \textit{natural language processing frameworks}, enabling the RAG system to retrieve more relevant documents.

With the advent of large language models (LLMs), natural language processing has seen unprecedented success across a range of tasks, owing to the impressive capability of LLMs to generate human language~\cite{anthropic2023model, https://doi.org/10.48550/arxiv.2210.11416, falcon40b, touvron2023llama, raffel2020exploring, brown2020language, chatGPT}. Despite these advancements, the application of LLMs for LEE remains largely unexplored.

This paper introduces LentEx, a novel, supervised learning framework specifically designed to address the challenges of LEE. LentEx leverages synthetic data generation and self-instruction to fine-tune smaller, efficient LLMs, thereby offering a scalable solution to LEE. By generating diverse training datasets that mirror the complexity of real-world text, LentEx establishes a robust foundation for extracting latent entities across multiple domains. These extracted entities serve as structured representations that improve text organization (clustering) and relevance filtering (retrieval-augmented generation). Our approach ensures generalization to unseen domains, making it highly applicable in real-world scenarios such as knowledge discovery, content personalization, and beyond. This work shifts the paradigm toward a deeper understanding of the implicit dimensions of text with the following key contributions:
\begin{enumerate}
    \item Formalizing domain-agnostic latent entity extraction problem and developing a systematic approach to address it.
    \item Developing a novel, template-based synthetic data generation process, producing diverse datasets without manually curated seed data.
    \item Demonstrating the effectiveness of fine-tuning a small LLM using this synthetic dataset, achieving superior performance over state-of-the-art embedding-based and generation-based methods on multiple downstream tasks.
\end{enumerate}

\section{Related Work}
\subsection{Latent Entity Extraction} Previous work in topic modeling has primarily centered around clustering textual data and identifying representative keywords to infer thematic patterns \cite{blei2003latent, miao2018discovering, grootendorst2022bertopic, Xu_2023_2}. These models typically require selecting an embedding space, followed by post-processing steps to extract interpretable topics from keyword clusters. While useful for high-level theme identification, such approaches often produce clusters of keywords that lack the granularity and specificity needed for actionable entity extraction~\cite{lample-etal-2016-neural, shoshan2018latent}. Traditional entity recognition and other information extraction tasks have relied heavily on supervised learning frameworks that map predefined labels to overtly mentioned entities \cite{lu2023pivoine}, thus limiting their adaptability to new, unseen categories.

Our approach overcomes these limitations by enabling the open-ended extraction of latent entities without reliance on predefined labels, advancing the field of entity extraction. By leveraging a more flexible framework, our method facilitates dynamic entity prediction even for out-of-distribution data, addressing the evolving nature of real-world datasets. This represents a leap forward in entity extraction, as it allows for adapting to new entity types and shifting data landscapes without retraining.

\subsection{Synthetic Data Generation} The high cost of curating and labeling data has long posed a challenge for training robust machine learning models~\cite{taori2023alpaca}. Recent advancements in synthetic data generation offer a promising solution, enabling the creation of large-scale, diverse training datasets at a fraction of the cost of manual annotation. Self-Instruct paradigm~\cite{wang2022self} demonstrates the power of prompting large language models to self-generate instructions, inputs, and outputs, resulting in labeled data points from limited seed data in few-shot setting.
 SeqGPT~\cite{yu2024seqgpt} collects data from multiple sources and leverages ChatGPT for data augmentation and label generation, while InstructPTS~\cite{fetahu2023instructpts} applies instruction-tuning techniques to generate product titles for retail domains.
To this end, our approach introduces a novel, template-based synthetic data generation process that produces diverse, contextually rich datasets without manually curated seed data. This method lowers the barrier to developing high-quality training data for latent entity extraction and contributes to the growing body of work focused on efficient, scalable synthetic data generation for NLP tasks.

\begin{figure}[t]
    \centering
    \includegraphics[width=0.99\linewidth]{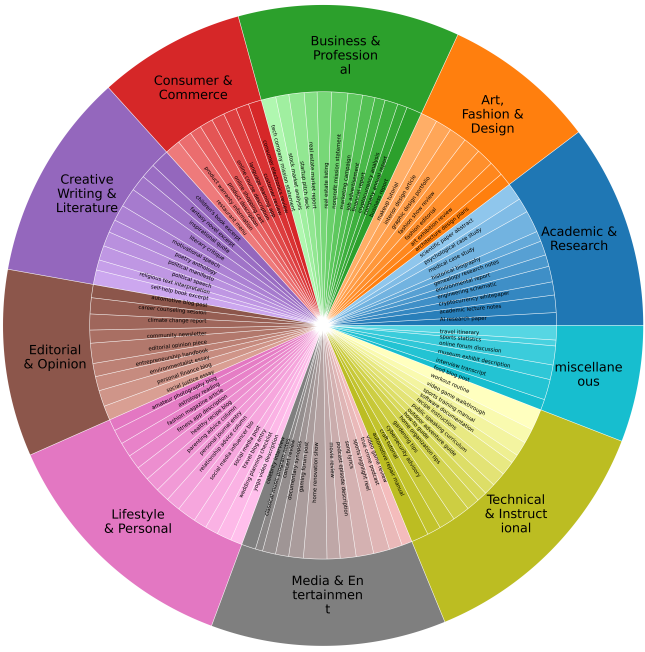} 
    
    \caption{Diversity of synthetic data. Inner circle represents the 100 synthetic domains in synthetic data. Outer circle represents the domains clustered based on their similarity}
    \label{fig:pie}
\end{figure}

\begin{figure}[t]
    \centering
    \includegraphics[width=0.99\linewidth]{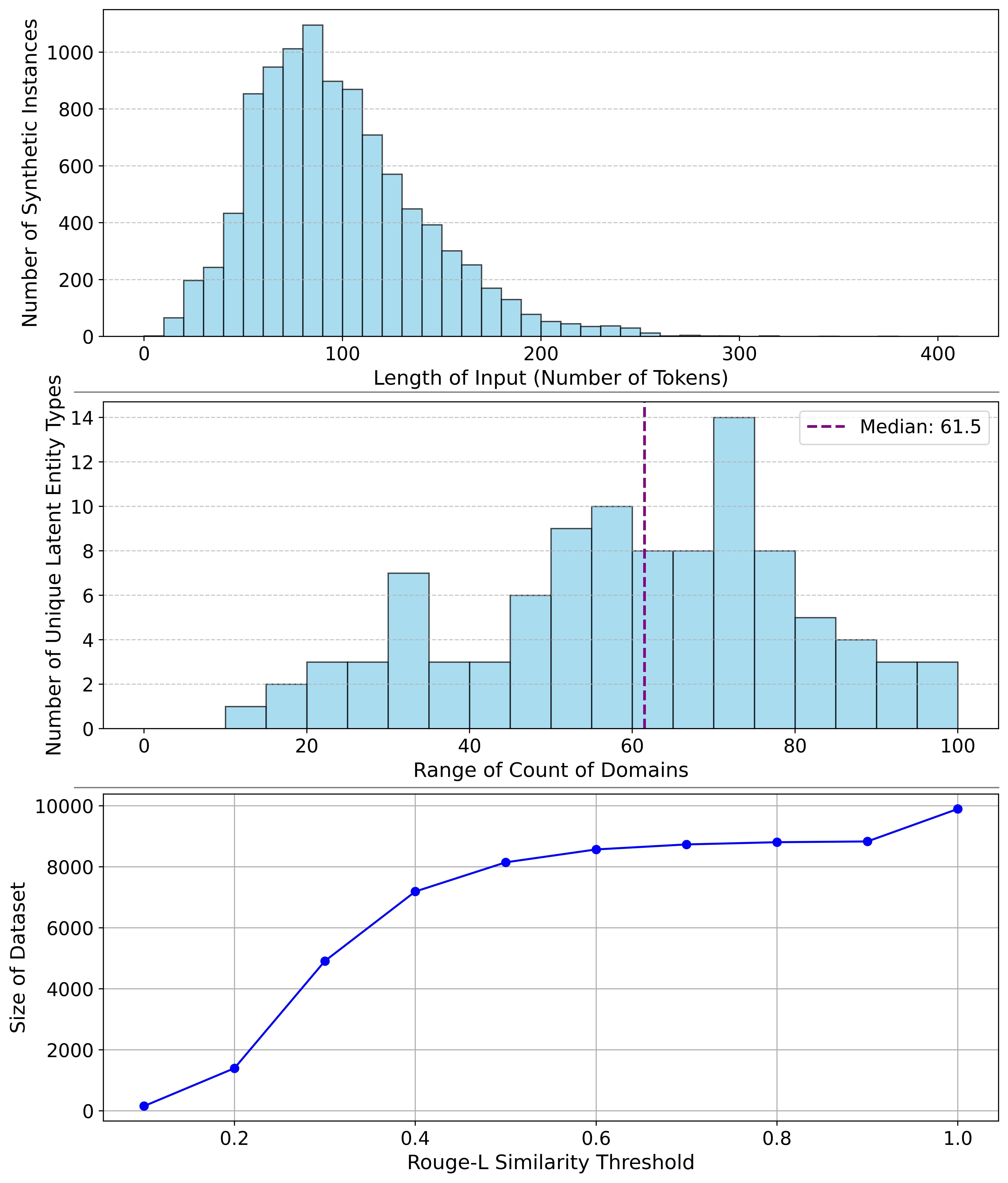} 
    
    \caption{Synthetic data distributions: Top figure shows the distribution of length of input tokens across instances, middle figure shows the distribution of range of count of domains and the corresponding number of unique entities, while the bottom figure shows the size of dataset at different values of Rouge-L}
    \label{fig:combined_three_plot}
\end{figure}

\section{LentEx}

\subsection{Problem Definition}
Latent Entity Extraction (LEE) involves identifying entities that are implied within the context of the text but are not explicitly mentioned. Defining formally:
\begin{equation}
    \label{eq:latent_entity_extraction}
    \text{E} = \mathcal{F}(\text{I}, \text{T}_{\text{entity}}, \text{T}_{\text{text}}, [\mathbf{O}])
\end{equation}

where $\mathcal{F}$ represents the function, i.e. the fine-tuned LLM, $\text{I}$ is the input text, $\text{E}$ denotes the identified latent entity ($\text{E} \not\in \text{I}$), $\text{T}_{\text{entity}}$ denotes the latent entity type (e.g., persona, industry), $\text{T}_{\text{text}}$ refers to the text type (e.g., narrative, description), and $\mathbf{[O]}$ is an optional set of predefined labels for supervised learning scenarios. This formulation enables LEE to function flexibly, addressing both unsupervised settings (e.g., topic modeling, clustering) and supervised tasks (e.g., classification).

\subsection{Method Overview}

LentEx addresses the challenge of labeled data scarcity through a novel methodology combining synthetic data generation and instruction fine-tuning, as illustrated in Figure~\ref{fig:example}. This approach enables effective latent entity extraction across multiple domains while using efficient, smaller LLMs.


\subsection{Synthetic Data Generation}


\begin{table*}[htbp]
\caption{
    Prompts for multiple stages. First row prompt is used in step 1, second row in step 2 of LentEx workflow. Third row shows the paraphrased prompts used in step 3 for instruction fine-tuning the model
}
\scriptsize
\centering
{
\begin{tabularx}{\textwidth}{p{3cm}X} 
    \toprule
    \textbf{Task} & \textbf{Prompt} \\
    \midrule
    \vspace{3em} Domain Generation \vspace{3em} & Generate a synthetic data dictionary containing 10 example entries. For each entry, the input should describe a broad category of text, and the output should specify a general category of latent entities. Latent entities are those that can be inferred from a text, even though they are not explicitly mentioned in the text. Ensure that the examples span a diverse range of topics, including but not limited to entertainment, education, technology, and health. The output categories should be broad, general, in 1-3 words, like 'movie genre' or 'product type', indicative of the kind of inference a reader can make from the given text type. Example: seed\_task\_dict = \{ "sports\ commentary": "sport\_name", "Google place review": "place category", "book summary": "book genre", "legal case\_summary": "legal case type" \} \\
    \midrule
    \vspace{3em} Input-output Instance Generation \vspace{3em} & Generate a list of 10 detailed instances, each with their corresponding type or category. For each instance, provide comprehensive content that is representative of its category, ensuring clarity, relevance, and authenticity. The instances should cover a diverse range of subjects, showcasing unique and specific characteristics of each category. Format the output as follows: Each instance should be encapsulated in a dictionary with two keys: "{text\_type}" and "{latent\_entity\_type}". The "{text\_type}" key should include the full content of the instance, detailed and structured appropriately for its type. The "{latent\_entity\_type}" key should specify the category or type that the instance belongs to. Structure each dictionary entry as a separate line in a JSONL format, where each line represents a single, complete dictionary corresponding to one instance and its category. Encapsulate the output in tag <rows>. \\
    \midrule
    \vspace{3em} Paraphrased Instructions for Training and Inference \vspace{3em} & \begin{enumerate}
    \item From the given \{text\_type\}, determine the \{latent\_entity\_type\}.
    \item Based on the provided \{text\_type\}, identify the \{latent\_entity\_type\}.
    \item From the specified \{text\_type\}, ascertain the \{latent\_entity\_type\}. 
    \item Using the given \{text\_type\}, pinpoint the \{latent\_entity\_type\}. 
    \item Examine the \{text\_type\} provided and determine the \{latent\_entity\_type\}. 
    \end{enumerate} \\
    \bottomrule
\end{tabularx}\par}

\label{tab:prompts}
\end{table*}

\subsubsection{Template-Based Domain Generation}~\label{sec:template}
Instead of manually curating seed dataset~\cite{wang2022self}, we adopt a template-based approach for synthetic data generation. We start with the prompt -  ``From the given \{text\_type\}, identify the \{latent\_entity\_type\},'' as shown in the prompts in Table \ref{tab:prompts} Using a few-shot learning setup, we leverage Claude 3 Sonnet~\cite{anthropic2023model} to iteratively generate $100$ unique domain combinations of {text\_type, latent\_entity\_type}, such as \textit{\{sport\_commentary: sport\_name\}}. This ensures a wide coverage of domain diversity. New combinations are generated at each iteration by selecting a random subset of previously created domains, and duplicates are filtered out. Figure \ref{fig:pie} illustrates the distribution of latent entity types clustered into broader parent domains.

\subsubsection{Synthetic Input-Output Instance Generation}

For each domain, we create synthetic input-output pairs using Claude 3 Sonnet. Each instance comprises a synthetically generated text and its corresponding latent entity label inferred from the context. This process is repeated over $10$ iterations per domain, generating a total of $10000$ diverse synthetic samples. The diversity of these samples is validated using Rouge-L similarity scores, with over 80\% of instances exhibiting scores below $0.5$, underscoring their variability. Figure~\ref{fig:combined_three_plot} shows the distribution of input token lengths, domain-entity mappings, and dataset sizes at different Rouge-L thresholds.


\subsection{Instruction Fine-Tuning}
\subsubsection{Training Data Construction}

We construct training data by concatenating instruction templates, domain descriptions, latent entity types, and input texts to form coherent model inputs, with the inferred latent entity as the target label. To ensure diversity and avoid explicit mentions of entities in the input text, we filter out instances with Rouge-L similarity scores above 0.5, forcing the model to rely on contextual inference. We mitigate over-fitting by employing paraphrased instruction templates (shown in Table~\ref{tab:prompts}, introducing variation in sequence structures, and applying subtle modifications in spacing and line breaks~\cite{chung2022scaling}.


\subsubsection{Model Selection and Training}
We fine-tune Mistral-7B-Instruct-v0.2~\cite{jiang2023mistral} using Low-Rank Adaptation (LoRA)~\cite{hu2021lora} technique for efficient training. 
The model is fine-tuned on four NVIDIA A10G GPUs for one epoch, with a batch size of $8$, a warmup ratio of $0.01$, and a learning rate of $0.0001$. LoRA parameters include rank $16$ and lora\_alpha $32$. Mistral-7B-Instruct was selected for its superior performance among models of similar size, and LoRA was chosen for its computational efficiency. Our methodology can be extended to other models and training setups.

\begin{table}[t]
\caption{
Experiment results from automated evaluation of model predictions on latent entity extraction task}
\centering
\small
\begin{tabularx}{\linewidth}{Xc} 
\toprule
\textbf{Model} & \textbf{Mean Semantic Similarity \%} \\
\midrule
Mistral-7B-Instruct-v0.2 & 61.17 \\
Claude-3-Haiku & 66.48 \\
LentEx (Ours) & 77.79 \\
\bottomrule
\end{tabularx}

\label{tab:semantic_sim}
\end{table}

\begin{table*}[t]
\caption{Clustering performance (v-measure scores) across different domains. p2p and s2s denote paragraph-to-paragraph and sentence-to-sentence tasks respectively. Best scores are in bold, second-best are underlined}
\centering
\renewcommand{\arraystretch}{1.2} 
\setlength{\tabcolsep}{7pt} 
\begin{tabular}{|p{2.8cm}|p{1cm}|p{0.5cm}|p{0.5cm}|p{0.5cm}|p{0.5cm}|p{0.5cm}|p{0.5cm}|p{0.5cm}|p{0.5cm}|p{0.5cm}|p{0.5cm}|p{1.2cm}|}
\hline
\multicolumn{13}{|c|}{\textbf{Inference without entity options (Clustering Task)}} \\

\hline

\textbf{Model} & \textbf{Average} & \multicolumn{2}{c|}{\textbf{arxiv-*}} & \multicolumn{2}{c|}{\textbf{biorxiv-*}} & \multicolumn{2}{c|}{\textbf{medrxiv-*}} & \multicolumn{2}{c|}{\textbf{reddit-*}} & \multicolumn{2}{c|}{\textbf{stackexchange-*}} & \textbf{twenty-newsgroups} \\

\cline{3-12}

 & & \textbf{p2p} & \textbf{s2s} & \textbf{p2p} & \textbf{s2s} & \textbf{p2p} & \textbf{s2s} & \textbf{} & \textbf{p2p} & \textbf{} & \textbf{p2p} & \\
\hline
gte-Qwen2-7B-instruct & 56.92 & \underline{56.46} & \underline{51.74} & 50.09 & 46.65 & 46.23 & 44.13 & \underline{73.55} & 74.13 & {79.86} & \underline{49.41} & 53.91 \\

bge-en-icl & 57.89 & 54.44 & 49.33 & \underline{53.05} & {48.38} & 45.86 & 44.33 & {72.33} & 72.72 & \underline{81.32} & {46.05} & \textbf{68.98} \\

NV-Embed-v2 & \underline{58.46} & 55.80 & {51.26} & \textbf{54.09} & \textbf{49.60} & \underline{46.09} & \underline{44.86} & 71.10 & \underline{74.94} & \textbf{82.10} & 48.36 & \underline{64.82} \\

\hline

Mistral-7B-Instruct-v0.2 & 39.74 & 40.18 & 37.30 & 36.94 & 35.16 & 40.74 & 39.53 & 48.10 & 36.58 & 45.81 & 35.41 & 41.44 \\
Claude 3 & 40.72 & 43.76 & 39.17 & 37.20 & 35.88 & 43.82 & 42.21 & 44.80 & 40.45 & 41.59 & 35.23 & 43.78 \\

\hline
LentEx (Ours) & \textbf{59.54} & \textbf{61.42} & \textbf{55.71} & 52.58 & \underline{48.96} & \textbf{49.15} & \textbf{46.85} & \textbf{74.63} & \textbf{79.59} & 64.90 & \textbf{56.63} & {64.56} \\
\hline
\end{tabular}
\label{tab:results2}
\end{table*}

\begin{table*}[htbp]

\caption{Retrieval Metrics for COLIEE and BioASQ Datasets. CAPTAIN results are only available for COLIEE as it is a legal domain-specific system}
\centering
\label{tab:combined_retrieval}
\begin{tabular}{lcccccccccc}
\toprule
\multirow{2}{*}{Model} & \multicolumn{3}{c}{COLIEE} & \multicolumn{3}{c}{BioASQ} \\
\cmidrule(lr){2-4} \cmidrule(lr){5-7}
                       & Precision & Recall & F-1 & Precision & Recall & F-1 \\
\midrule

RAG (Baseline) & 0.248 & 0.167 & 0.199 & 0.612	&	0.441	&	0.513 \\
RAG + Mistral & 0.644 & 0.536 & 0.585 & 0.681	&	0.496	&	0.574 \\
RAG + Claude & 0.715 & 0.663 & 0.688 & \underline{0.704}	&	\underline{0.528}	&	\underline{0.603} \\
RAG + Pre-trained NER & 0.652 & 0.513 & 0.574 & 0.657	&	0.513	&	0.576 \\
RAG + Topic Modeling & 0.667 & 0.591 & 0.627 & 0.649	&	0.502	&	0.566 \\
CAPTAIN & \underline{0.787} & \textbf{0.708} & \textbf{0.745} & -	&	-	&	- \\
\textbf{RAG + LentEx (Ours)}  & \textbf{0.792} & \underline{0.681} & \underline{0.732} & \textbf{0.742}	&	\textbf{0.595}	&	\textbf{0.66} \\

\bottomrule
\end{tabular}
\end{table*}

\section{Experimentation}~\label{sec:exp}
We perform a series of experiments to evaluate the effectiveness of LentEx on latent entity extraction as well as its downstream applications in Clustering and Retrieval-Augmented Generation (RAG).

\subsection{\textbf{Latent Entity Extraction Performance}}
In this experiment, we evaluate LentEx's ability to extract latent entities accurately.
We evaluate the LentEx framework by comparing predicted latent entities against reference labels generated from synthetic data. We compute the semantic similarity between model predictions and synthetic labels using the all-mpnet-base-v2~\cite{song2020mpnet} embedding model. Mean semantic similarity is calculated for all instances to quantify the alignment between predicted and reference entities.
We compare LentEx against two baselines: Mistral-7B-Instruct-v0.2~\cite{jiang2023mistral} and Claude-3-Haiku~\cite{anthropic2023model}. We randomly sample $50$ synthetic instances from each domain, totaling $5000$ instances for evaluation.

\textbf{Result: }As shown in Table~\ref{tab:semantic_sim}, LentEx achieves higher semantic similarity with the reference labels, compared to Mistral-7B-Instruct-v0.2 and Claude-3-Haiku, establishing the baseline effectiveness of LentEx as an entity extraction model.

\subsection{\textbf{Clustering with Latent Entity Representations}}

To demonstrate the effectiveness of LentEx in real-world scenarios, we apply LentEx to clustering tasks using $11$ P2P (paragraph input) and S2S (sentence input) datasets from the Multilingual Textual Entailment Benchmark (MTEB)~\cite{muennighoff2022mteb}. These datasets span diverse domains, including arXiv, bioRxiv, medRxiv, Reddit, StackExchange, and 20-Newsgroups. We benchmark LentEx against state-of-the-art embedding-based models: gte-Qwen2-7B-instruct~\cite{qwen2}, bge-en-icl~\cite{bge_embedding}, NV-Embed-v2~\cite{lee2025nvembedimprovedtechniquestraining} (as of Oct 2024), as well as prompting based methods: Mistral-7B-Instruct-v0.2~\cite{jiang2023mistral} and Claude-3-Haiku~\cite{anthropic2023model}.

For all generation methods including ours, we conduct zero-shot inference to infer the latent entity. We embed the latent entities, followed by a mini-batch k-means model with batch size $500$ and $k$ equal to the number of different labels~\cite{muennighoff-etal-2023-mteb}. We choose the metric in accordance with MTEB, i.e., v-measure \cite{rosenberg-hirschberg-2007-v} for Clustering, which is computed as the harmonic mean of distinct homogeneity and completeness scores.
\begin{equation}
    \label{eq:v_measure}
    V_{\beta} = (1 + \beta) \cdot \frac{H \cdot C}{(\beta \cdot H) + C}
\end{equation}
where \textbf{H} is the homogeneity score, \textbf{C} is the completeness score, \(\beta\) is a parameter that balances the trade-off between homogeneity and completeness. If \(\beta = 1\), then the V-measure is the harmonic mean of homogeneity and completeness.

\textbf{Result:} 
Experimental results in Table~\ref{tab:results2} demonstrate that LentEx consistently outperforms both embedding-based and prompting-based baselines across most datasets. LentEx achieves the highest average v-measure score of $59.54$, outperforming strong embedding-based baselines like NV-Embed-v2 ($58.46$) and bge-en-icl ($57.89$). The performance gains were particularly pronounced in paragraph-level tasks, with notable improvements in arxiv-p2p ($61.42$ vs. $56.46$) and reddit-p2p ($79.59$ vs. $74.94$).
These results demonstrate LentEx's robustness and generalizability across different text granularities (p2p and s2s) and domain types, from scientific literature to social media content.

\subsection{\textbf{Improving Retrieval in RAG with Latent Entity Filtering}}

While clustering demonstrates LentEx's information structuring capabilities, we further evaluate it in retrieval tasks by integrating it into a RAG pipeline.



\subsubsection{\textbf{Datasets}}
We conduct our evaluation  on two distinct domain-specific datasets: COLIEE~\cite{coliee2018} for legal domain and BioASQ~\cite{bioasq2015} for biomedical domain. The COLIEE test set comprises $100$ legal queries and $375$ supporting documents.
We identify the latent entity types for this dataset including \textit{case\_type, parties\_involved, jurisdiction, case\_outcome}.
For the BioASQ-2023 \textit{11b Phase A} dataset~\cite{BioASQ_2023}, we utilize the Summary questions from training set, which consists of $1130$ questions and $4719$ supporting documents.
The latent entities extracted for this domain are specifically selected to capture key biomedical concepts: \textit{diseases, drugs, symptoms, treatments, anatomical\_terms, and biomarkers}. These latent entities serve as crucial pre-filters for enhancing the relevance of evidence retrieval.

\subsubsection{\textbf{Baselines}}
We compare our RAG+LentEx system against a suite of baselines to assess the effectiveness of latent entity filtering. The primary baseline implemented a standard RAG architecture~\cite{lewis_rag} performing retrieval without latent entity pre-filtering. We further compare against four RAG variants incorporating named and latent entity extraction methodologies: (i) RAG+Mistral, utilizing Mistral-7B-Instruct-v0.2~\cite{jiang2023mistral}, (ii) RAG+Claude, utilizing Claude-3-Haiku~\cite{anthropic2023model} for latent entity extraction, (iii) RAG+NER: RAG augmented with a pre-trained Named Entity Recognition (NER) model~\cite{lample_ner}, and (iv) RAG+Topic Modeling, which implemented topic-based entity pre-filtering~\cite{blei2003latent}. All systems maintain identical retrieval and embedding configurations, including paragraph-level chunking and dual-encoder architecture with Qwen1.5-7B-instruct~\cite{li2023towards}.

\begin{figure}[t]
    \centering
    \includegraphics[width=0.99\linewidth]{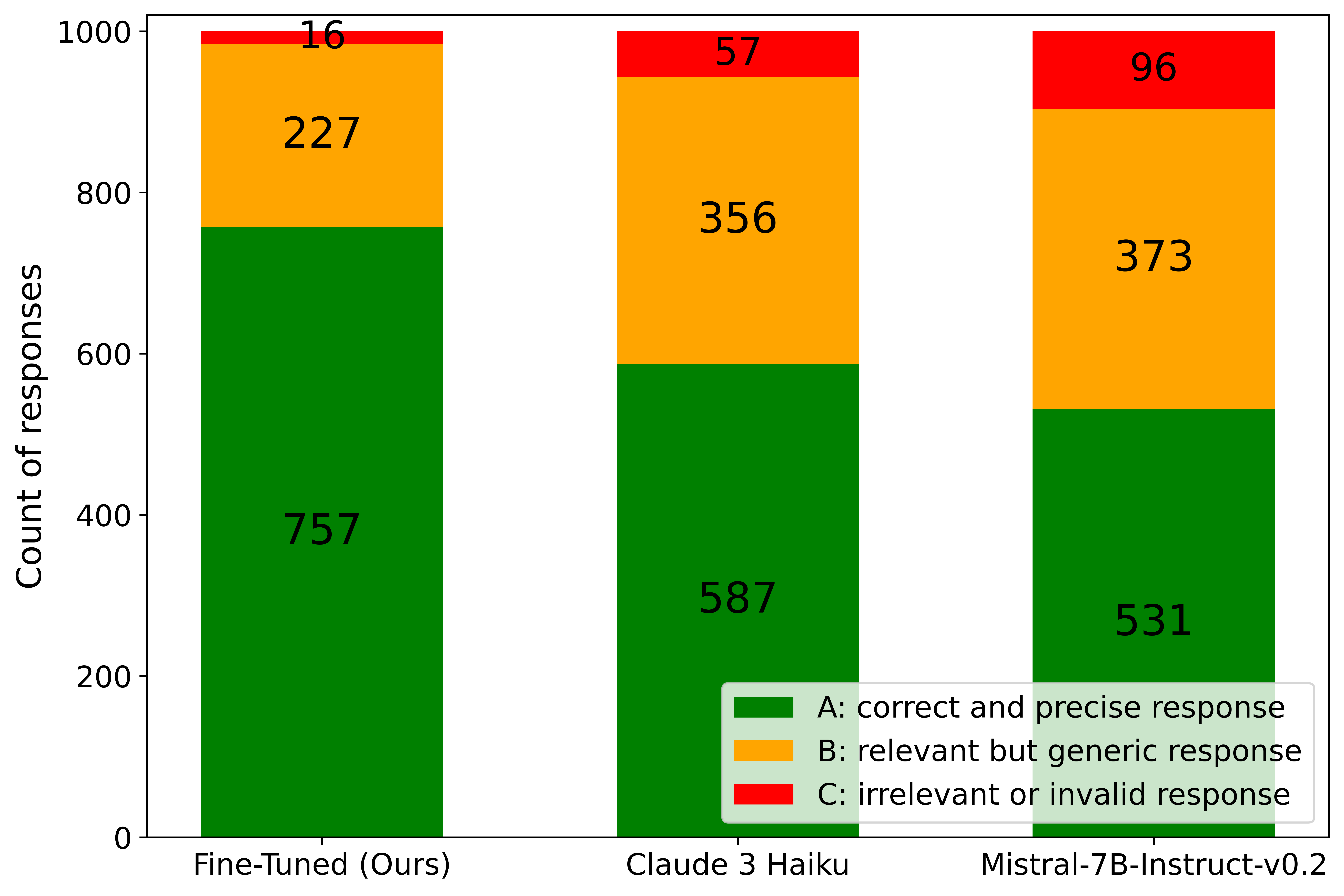} 
    
    \caption{LLM evaluation of model responses over $1000$ data points. Claude-3-Sonnet model was prompted to rate model responses as A, B, or C based on the SOP shown. Evaluations for different models were conducted independently, allowing for identical ratings across models.}
    \label{fig:performance}
\end{figure}

\subsubsection{\textbf{Experimental Setup}}

We segment documents into paragraph-level chunks (max length: $512$ tokens, overlap: $10$ tokens) for fine-grained retrieval, embedding them using Langchain's OpenSearch with gte-Qwen1.5-7B-instruct~\cite{li2023towards}. LentEx extracts latent entities from paragraphs and stores them as metadata. During inference, LentEx extracts entities from queries in real-time and filters documents based on metadata alignment before retrieval to enhance result relevance. 
Initial observations show that excessive filtering parameters excludes relevant documents, while insufficient filtering yields non-relevant results. The heterogeneous nature of entity types across diverse questions presented challenges in developing a universally applicable query protocol. To address these limitations, we implement dynamic queries using OpenSearch Query-DSL~\cite{opensearch2024} with varying entity combinations as filters. This approach leverages OpenSearch's \textit{min\_should\_match} parameter to specify the required quantity of matching filters, iteratively adjusting query restrictiveness. This facilitates query-specific configuration of both the combination and quantity of filters necessary for optimal alignment between question and document metadata. Following retrieval, we apply ms-marco-MiniLM-L-6-v2~\cite{minilm} for re-ranking, extracting the top $10$ results.

\textbf{Results}
 The experimental results in Table~\ref{tab:combined_retrieval} demonstrate the effectiveness of our RAG+LentEx approach across both legal and biomedical domains.
On the COLIEE legal dataset, RAG+LentEX achieved the highest precision ($0.792$) and competitive recall ($0.681$), resulting in an F1-score of $0.732$. This represents a significant improvement over the baseline RAG system, which achieves only $0.199$ F1-score. Notably, our system's performance is comparable to CAPTAIN~\cite{captain}, the current state-of-the-art system, even slightly outperforming it in precision ($0.792$ vs $0.787$).
In the biomedical domain (BioASQ), RAG+LentEx achieves the highest scores across all metrics (P: $0.742$, R: $0.595$, F1: $0.66$). The second-best performance is achieved by RAG+Claude (F1: 0.603), suggesting the effectiveness of large language models in entity extraction for specialized domains. The performance of LentEx across both domains demonstrates its robustness in diverse domain-specific retrieval tasks.

\begin{figure}[t]
    \centering
    \includegraphics[width=0.99\linewidth]{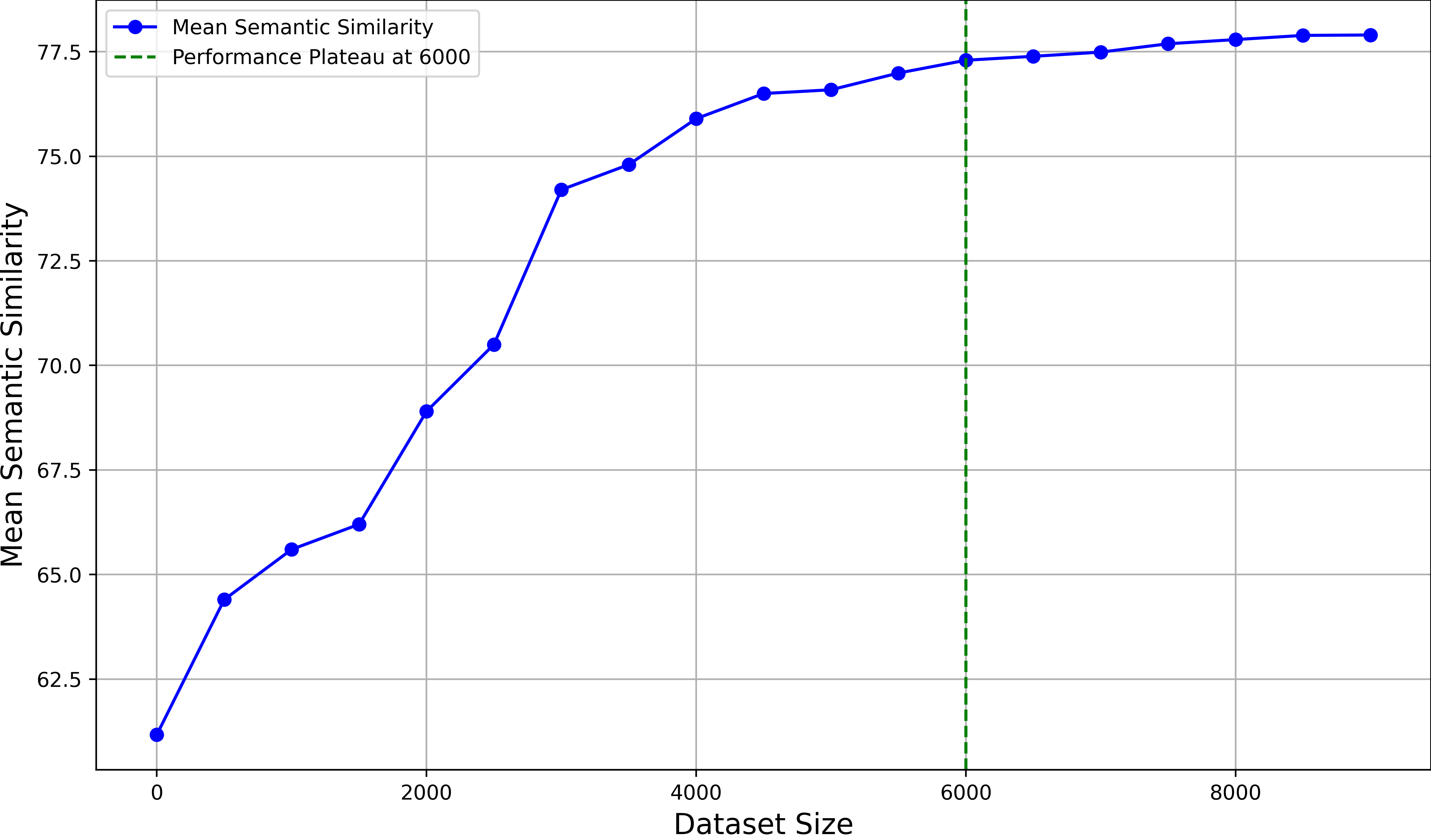} 
    
    \caption{Effect of dataset size on model performance. Using synthetic outputs as reference labels, semantic similarity is measured between model predictions and labels. The y-axis represents the mean semantic similarity on test dataset.}
    \label{fig:dataset_size}
\end{figure}

\subsection{\textbf{LLM as an Evaluator}}

Next, we assess the relevance of model outputs by categorizing them into three buckets: A representing correct and precise response, B representing relevant but generic response, and C representing irrelevant or invalid response. We use Claude-3.5-Sonnet~\cite{anthropic2023claude35} as the evaluator model~\cite{evaluator}, which classifies the output based on these categories. The order of model outputs is shuffled for each evaluation to ensure unbiased assessments.

\textbf{Result: }As depicted in  Figure~\ref{fig:performance}, LentEx significantly outperforms both baseline models, with 98.4\% of instances classified as either A or B, confirming the robustness of its fine-tuning approach.

\section{Ablation Studies}
\subsection{\textbf{Impact of Dataset Size on Model Fine-Tuning}}
We systematically investigate the effect of varying dataset sizes on model fine-tuning to determine the optimal amount of data required for efficient training. The model is fine-tuned using progressively larger subsets of training data, and its performance is evaluated on a held-out test set. As depicted in
Figure~\ref{fig:dataset_size}, the model’s performance exhibits consistent improvement up to a dataset size of $6000$. Beyond this point, the performance plateaus, suggesting that $6000$ samples constitute a sufficient data size for achieving optimal fine-tuning outcomes under the current setup. This result highlights the model’s efficiency in learning from a moderate-sized dataset without extensive data collection.

\subsection{\textbf{Effect of Synthetic Data Generation on Data Quality}}
To evaluate the effectiveness of the template-based synthetic data generation strategy introduced in Section~\ref{sec:template}, we conduct a manual assessment on $2000$ randomly sampled examples, with $20$ instances drawn from each domain. The evaluation focuses on two critical aspects: (1) the alignment between the input text and its designated domain and (2) the accuracy of the output latent entity labels. As shown in Table~\ref{tab:data_quality_eval}, $98\%$ of the sampled instances demonstrate alignment with their respective text types, and $91.5\%$ of the output labels are deemed correct. In contrast, when the output labels are generated independently after all input instances are created \cite{wang2022self}, the correctness rate drops to $91\%$. These results affirm that generating input-output pairs within the same model call leads to higher data quality and reinforces the efficacy of our template-based synthetic data generation pipeline.

\begin{table}[t]
\caption{Manual data quality evaluation of input texts and labels from synthetic data, on a stratified sample of 2000 examples}
\centering
\small
\begin{tabularx}{\linewidth}{Xc} 
\toprule
\textbf{Quality Review} & \textbf{Yes \%} \\
\midrule
Is synthetic input text aligned with the text type? & 98 \\
Is latent entity label correct? & 91.5 \\
\bottomrule
\end{tabularx}
\label{tab:data_quality_eval}
\end{table}

\section{Results and Discussion}

LentEx excels in addressing latent entity extraction (LEE) through a practical and cost-effective approach that leverages synthetic data generation and instruction fine-tuning. This methodology is particularly well-suited for low-resource environments where labeled data is scarce, enabling the model to generalize across a wide range of domains. Importantly, while LentEx is trained exclusively on synthetic data, our comprehensive evaluation on diverse real-world datasets—including 11 MTEB clustering benchmarks spanning scientific literature, social media, and news, along with specialized legal (COLIEE) and biomedical (BioASQ) retrieval datasets—demonstrates robust generalization capabilities. The strong performance across varied datasets validates that our synthetic data generation successfully captures the complexity and distribution of naturally occurring text, enabling effective transfer to practical applications without domain-specific labeling. The model's versatility is further evidenced by its performance in zero-shot settings, while its integration into Retrieval-Augmented Generation (RAG) systems boosts retrieval relevance and precision. LentEx's optimal performance with moderate dataset size underscores its efficiency in balancing data quality and diversity, making it a highly adaptable tool for both academic research and industry applications where latent entity recognition enhances content personalization, knowledge discovery, and information retrieval systems.


\section{Error Analysis}

Analysis of LentEx on clustering and RAG experiments revealed key insights into it's performance and limitations. In clustering, errors primarily occurred in ambiguous entity assignments, where multiple latent entities existed but LentEx selects a suboptimal one. This effect is more profound in datasets such as twenty-newsgroups and stackexchange, where the type of latent entity is generic. This leads to mis-classified clusters, leading to inferior performance. In RAG experiment, errors are linked to over-restrictive filtering, where relevant documents are excluded due to incomplete latent entity extraction. This is particularly noticable in COLIEE dataset, where case law often involves intersecting legal categories (e.g., \textit{contract law vs. tort law}), and LentEx may miss documents if it fails to infer all relevant latent entities from the query. Addressing these limitations through domain-specific fine-tuning and adaptive entity selection strategies could further enhance LentEx's performance in real-world applications.

\section{Conclusion and Future Work}

We present LentEx, a novel, supervised framework for latent entity extraction, leveraging synthetic data generation and instruction fine-tuning to overcome the challenge of labeled data scarcity. LentEx demonstrates superior generalization across multiple domains and tasks, outperforming state-of-the-art models. Our experiments validate the efficacy of LentEx on various downstream tasks, including clustering and retrieval-augmented generation, which enhances the quality and relevance of retrieved results.

While LentEx sets a strong foundation for latent entity extraction, several avenues remain for future research. First, further refinement of filtering mechanisms in retrieval-augmented systems is needed to address the observed drop in Recall. Incorporating adaptive filtering strategies, possibly informed by context-aware retrieval, could mitigate this limitation. Although our synthetic data generation strategy has proven effective, exploring more sophisticated techniques, such as self-supervised training, contrastive learning or reinforcement learning, could further enhance the model performance. Another critical direction is the application of LentEx to more complex, domain-specific tasks, such as temporal reasoning or multi-hop inference, where latent entities can play a pivotal role in understanding and decision-making. LEE provides richer, structured representations that can be integrated into knowledge graphs, search engines, and content recommendation systems. Lastly, addressing ethical concerns and potential biases inherent in large-scale language models will be crucial. We highly encourage researchers to build on LentEx by exploring training strategies, expanding generalization capabilities, and addressing ethical considerations.


\bibliographystyle{IEEEtran}
\bibliography{ijcnn}

\end{document}